\documentclass[twocolumn]{article}
\usepackage{cite}
\usepackage{amsmath,amssymb,amsfonts}
\usepackage{algorithmic}
\usepackage{graphicx}
\usepackage{textcomp}
\usepackage{multirow}
\usepackage{abstract}
\usepackage{url}

\graphicspath{{figures/}}

\begin{document}

\title{A Fast Hybrid Cascade Network for Voxel-based 3D Object Classification}
\author{Ji Luo$^{1}$, Hui Cao$^{2}$, Jie Wang$^{3}$, Siyu Zhang$^{2}$ and Shen Cai$^{2,*}$ \\
$^{1}$School of Management Science and Engineering, \\ Shandong University of Finance and Economics \\
$^{2}${Visual and
Geometric Perception Lab, Donghua University} \\
$^{3}${Department of Computer Science, University of Manchester} \\
$^{*}$Corresponding author: Shen Cai (e-mail: hammer\_cai@163.com). }

\twocolumn[
\begin{@twocolumnfalse}
\maketitle 
\begin{abstract}
Voxel-based 3D object classification has been thoroughly studied in recent years. Most previous methods convert the classic 2D convolution into a 3D form that will be further applied to objects with binary voxel representation for classification. However, the binary voxel representation is not very effective for 3D convolution in many cases. 
    In this paper, we propose a hybrid cascade architecture for voxel-based 3D object classification. It consists of three stages composed of fully connected and convolutional layers, dealing with easy, moderate, and hard 3D models respectively. Both accuracy and speed can be balanced in our proposed method. By giving each voxel a signed distance value, an obvious gain regarding the accuracy can be observed. Besides, the mean inference time can be speeded up hugely compared with the state-of-the-art point cloud and voxel based methods.
\end{abstract}
\end{@twocolumnfalse}
~\\
]


\vspace{50pt}
\section{Introduction}

In the past several years, convolutional neural networks (CNNs) have performed well in many computer vision tasks, such as object classification, detection, and segmentation. Three-dimensional (3D) object recognition is very important when it comes to 3D environment understanding, which can be used in a wide range of applications like self-driving vehicles and autonomous robots. Compared to two-dimensional (2D) images captured by RGB cameras, 3D data is usually acquired by depth cameras or lidar sensors, which provide richer geometrical information in three-dimensional space. Therefore, shape is the most important property when handling 3D data, while color, texture and illumination are somewhat more concerned in 2D image processing.  

Traditional methods usually utilize 3D hand-crafted features to describe 3D objects \cite{rusu2009fast}, and support vector machine (SVM) is usually applied for classification based on the extracted 3D shape features. However, the performance is quite limited. With the rapid development of deep learning, it has been proven to be effective by incorporating the deep features in many fields. Deep learning based approaches have also been adopted to improve the classification performance of a certain 3D object representation.
\begin{figure*}[th]
\setlength{\belowcaptionskip}{-15pt}
\begin{center}
\includegraphics[width=1.0\textwidth]{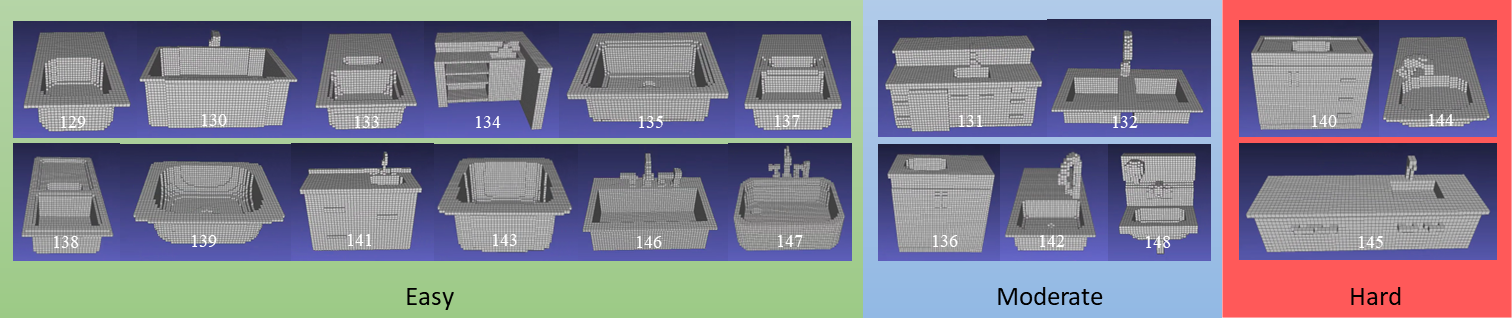}
\vspace{-12pt}
\caption{Models of `sink' class (with the serial number below) in ModelNet40~\cite{ModelNet40} test set classified through three stages of our cascade network. For example, there are $12$ `Easy' models ($60\%$ of all $20$ tested models) classified by the first stage of the cascade network.
`Easy', `Moderate', and `Hard' categories are visually consistent with human sense.}
\vspace{4pt}
\label{fig:sink_example}
\end{center}
\end{figure*}

A pioneering work of MVCNN~\cite{Su2015MultiviewCN} attempts to apply 2D CNN on 3D shape analysis, which first renders a 3D object from several different viewpoints and extracts features from those 2D projections with 2D CNN feature extractor separately. Its variants include MVCNN-new~\cite{su2018a}, RotationNet~\cite{kanezaki2018rotationnet:}, and etc. 
These image-based methods benefit from the existing well-developed network architectures, such as VGG~\cite{VGG} and ResNet~\cite{He2016DeepRL}, and achieve a good performance in 3D object classification. 
However, these approaches are not geometrically intuitive and cannot easily be extended to other 3D tasks, such as part segmentation. 
 
Instead of using 2D CNN, some methods directly handle the surface point cloud of a 3D object. PointNet~\cite{Qi2017PointNetDL} processes point clouds with a simple yet efficient multi-layer perceptron (MLP) network. However, it only considers global features and ignores local neighborhood information, making it not suitable to fine-grained shape and complex scenes. Instead of working on individual points, PointNet++~\cite{Qi2017PointNetDH} introduces a hierarchical neural network that applies PointNet recursively on the group points in different levels.
To better utilize local geometric structure,
DGCNN~\cite{Wang2018DynamicGC} constructs a local neighborhood graph and applies edge convolutions on the connecting points to achieve a high accuracy in classification.

Similar to point-based methods, volumetric CNNs can directly process 3D data. The difference is that volumetric methods first transfer an object to volumetric occupancy representation and then conduct 3D convolution on the voxel occupancy data. The early volumetric 3D CNNs are introduced in ShapeNet~\cite{Wu20153DSA} and VoxNet~\cite{Maturana2015VoxNetA3}. These methods achieve acceptable classification accuracy but the performance is highly restricted by the low-resolution representation due to high memory requirements and computation cost. To handle these problems, octree-based CNNs~\cite{wang2017o-cnn,yuan2018ocnet,klokov2017escape} propose to represent the object with a sparse grid-octree structure and implement 3D CNN on the octree grid voxel. 
Ma et al. \cite{ma2019binary} introduces binary weights in volumetric CNNs, which can reduce the calculation cost and accelerate the network by efficient bit-wise operations. Prokudin et al.~\cite{prokudin2019efficient} proposes
several kinds of basis point sets (BPS) to efficiently reconstruct and classify 3D objects. The main problem of voxel-based methods is that the classification accuracy is much lower than that of surface point cloud. 

In summary, it is quite important to lift the classification accuracy, while the running speed should also be ensured. In this paper, we try to balance the two aspects and carry out the 3D object classification. Overall, the contributions of our work can be concluded as follows:

(1) We propose to directly use the signed distance field (SDF) voxel for both fully connected network (FC-Net) and 3D CNNs. The SDF voxel representation promotes the classification performance of single network greatly. 

(2) A hybrid three-stage cascade architecture combining FC-Net with 3D CNN for voxel-based object classification is proposed. It improves the classification accuracy and decreases the mean inference time significantly.

(3) We propose to use an adaptive thresholding strategy with respect to confidence of each class in the cascade network to achieve a high pass rate and a high precision simultaneously. 

The visual classification results of `sink’ models are shown in Fig.~\ref{fig:sink_example} to illustrate the `Easy', `Moderate', and `Hard' shapes classified by the three stages of the proposed cascade network.

\section{Related Works}
In this section, we will briefly review the previous works closely related to our work, including implicit representations of 3D shape, voxel-based 3D CNNs and cascade classifiers.

\subsection{Implicit Representations}
In contrast to explicit representations which encode the shape discretely (e.g. voxels and surface points), implicit representations encode shapes continuously. Occupancy networks~\cite{ chen2019learning, mescheder2019occupancy} implicitly represent the 3D
surface as the continuous decision boundary with a deep neural network classifier by inferring an occupancy probability. SDF is another popular choice for 3D shape representation. Voxel-based SDF have been extensively used for 3D shape representations~\cite{stutz2018learning,park2019deepsdf, michalkiewicz2019implicit}. In 3D object classification, Cao et al.~\cite{cao2020inspherenet:} propose to construct a class of more representative features named infilling spheres with SDF values as their radii and it achieves better accuracy than PointNet~\cite{Qi2017PointNetDL} and PointNet++~\cite{Qi2017PointNetDH}.

\subsection{Voxel based 3D CNNs}
Early works of volumetric CNN ~\cite{maturana2015voxnet, Wu20153DSA} propose to convert 3D shape to binary representation on voxel grids, and then apply a 3D convolution network to extract the global feature for shape analysis. However, they are unable to handle large scale scenes since the computation and memory cost grows greatly with the cube of the resolution increases. To address the limitation of high computation and memory cost, the works~\cite{ma2019binary, 3DSparseConvolutional2021, 3DVoxelCapsule2020, CAI2021} transform the inputs and parameters into binary values which largely accelerates the neural network by bit-wise operations. Another family of methods~\cite{wang2017o-cnn, yuan2018ocnet, klokov2017escape, VoxTr2021, Wang-2022-dualocnn} falls in the volumetric shape with octree or sparse structure, which require less computation and memory cost but can still achieve a good accuracy in shape analysis tasks. Prokudin et al. \cite{prokudin2019efficient} proposes to encode point clouds with a fixed-length feature vector of basis point sets (BPS), which can be computed efficiently with several standard neural network architectures. The accuracy of BPS based approach is pretty high among all existing voxel-based methods. 
Some volumetric methods are fast and avoid high memory and computation cost \cite{maturana2015voxnet, real-time}, but they are not able to achieve a relatively high recognition accuracy.  In general, the 3D convolution operations in volumetric CNNs usually take a lot of computational consumption. The number of parameters evolved in the fully connected (FC) layers can be huge, leading to great memory and storage.
\begin{figure*}[tbp]
\begin{center}
\includegraphics[width=0.99\textwidth]{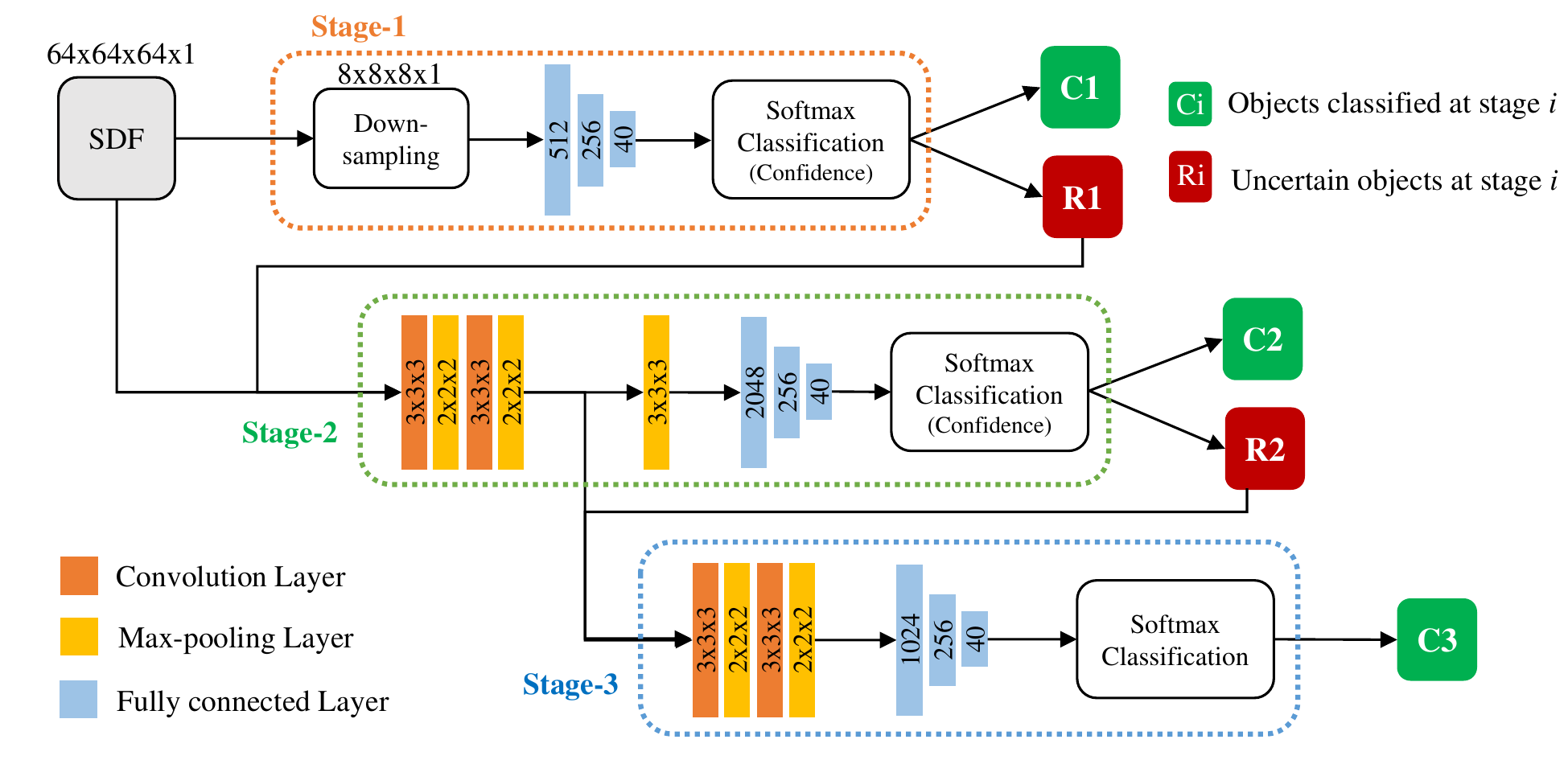}
\vspace{-8pt}
\caption{Hybrid cascade network architecture for SDF voxel classification. `C1', `C2', and `C3' passing through the different stage of our cascade network correspond to `easy', `moderate', and `hard' categories, respectively.}\label{fig:cascade}
\end{center}
\vspace{-8pt}
\end{figure*}

\subsection{Cascade Classifiers}
Cascade classifier has been used in traditional object detection tasks, such as the classical Haar cascade classifer in face detection~\cite{viola2001rapid}.
Cascade strategy is also introduced in object detection neural networks~\cite{ouyang2017chained, shivajirao2019mask}, which train a sequence of detectors stage by stage with increasing intersection over union (IoU) thresholds. Cascade detection networks often reject easy negative samples like background at early stages for better detection and faster inference. Since a large number of obvious negative samples are discarded at early stages, it reduces the computational and memory cost because those discarded samples are no longer paticipating the calculation in deeper layers. Huang et al. \cite{huang2017multi-scale} uses a multi-scale network architecture and trains multiple classifiers with varying resource demands to improve the average accuracy.
For semantic segmentation task, Li et al. \cite{li2017not} proposes a deep layer cascade method, in which earlier convolution layers are trained to handle easy regions, while the hard regions are fed to the deep layers for segmentation.

\section{Hybrid Cascade Network}

To find a balance between high accuracy and fast inference speed, we adopt the idea of cascade network and propose a hybrid network based on fully connected and convolutional layers for voxel-based 3D object classification.
The detailed hybrid cascade architecture is depicted in Figure \ref{fig:cascade}, which consists of three main parts. The classifier in the first stage is a two-layer fully connected network (FC-Net). The second classifier is a shallow 3D CNN, while the third classifier is a deep 3D CNN.

Since three classifiers are cascaded, the whole workflow works through the following pipelines. If a sample's confidence score of the top-1 category is high enough, then it will be classified in the current stage. Otherwise, it will be considered as an uncertain shape and sent to the next classifier. An adaptive threshold strategy is applied to determine the condition. As a result, `C1', `C2', and `C3' are normally corresponding to `easy', `moderate', and `hard' categories, respectively.

\subsection{Input Signal}

It is worth noting that the input signal we adopt is the signed distance field of voxel, rather than binary voxel. 

Most volumetric CNN methods take binary signal as input and learn a probability distribution of binary variables on voxel grids, where ``1" stands for object occupancy and ``0" indicates the empty. Although the encouraging performance has been achieved, the accuracy of almost all methods are still lower than $90\%$. One basic factor leading to low classification accuracy is that binary volumetric representation could not provide detailed shape features for deep learning. In Sec.~\ref{sec:comparison}, we perform several comparison experiments between different input signals. It is found that SDF values of voxels behave much better than their binary features in FC-Net and 3D CNN. Therefore, our further research mainly take SDF feature as input to improve the performance. The original voxel resolutions we use in this paper are $16^3$, $32^3$, $64^3$.

\begin{figure}
\setlength{\belowcaptionskip}{-10pt}
\begin{center}
\includegraphics[width=0.48\textwidth]{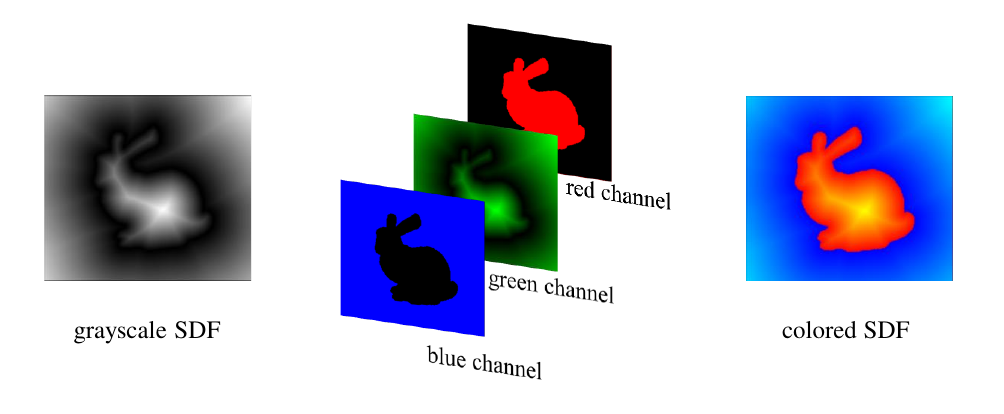}
\vspace{-12pt}
\caption{Colored SDF of 2D Bunny. The colored SDF is actually a hybrid multi-channel representation, which takes into account the edge information (implied in the binary fields of the red and blue channels), as well as the changes inside and outside the shape (implied in the grayscale SDF of the green channel).}\label{fig:color_SDF}
\end{center}
\vspace{-12pt}
\end{figure}

Besides the vanilla SDF input, we also propose to use colored SDF feature to further enrich the input, which is inspired by the pseudo-colorization method of gray images. Figure~\ref{fig:color_SDF} depicts one example of Bunny rabbit represented by colored SDF. 
The grayscale SDF is assigned to the green channel, while the binary field is assigned to the red and blue channels.
As a result, colored SDF can be considered as a hybrid multi-channel representation, which takes into account the edge information and the changes inside and outside the shape.
From a human point of view, colored SDF are more illustrative and may contain more prominent and easy-to-extract features. 

\subsection{FC-Net based Stage}
Although the nonlinear fitting ability of FC-Net with more than two hidden layers is very strong, it often leads to over-fitting and slow reference speed. 
We empirically find that a satisfactory classification accuracy can be achieved with a simple FC-Net consisting of only two hidden layers, whether it is used alone or in the first stage of the cascade classifier. Moreover, the voxel input of FC-Net only needs to be down-sampled to $8\!*\!8\!*\!8$ resolution at the equal intervals Thus, the FC-Net in the first stage of our cascade network not only is very small, but also can runs very fast.


\subsection{3D CNN based Stages}

As illustrated in Figure \ref{fig:cascade}, if the top-1 score of FC-Net output at stage one is lower than the threshold, the voxel with the original resolution of the sample will be alternatively fed to the second classifier network which consists of two convolution blocks and a classification head similar to FC-Net. Specifically, the operation sequence ``3D convolution + batch normalization (BN) + ReLU + max pooling” is adopted as a basic block. After the operation of two blocks, a max pooling with size of 3 is performed to further down sample the feature volume from 14*14*14*32 to 4*4*4*32. The the down-sampled feature is sent to the classification head of the second stage. Similarly, the comparison between the top-1 score and the threshold is performed again at the second stage. If the score is lower than the threshold, the feature generated through the convolution blocks in the second stage will go through the third classifier. The third stage contains another two convolution blocks and a classification head. Compared to the second stage, the main difference in third stage is that no max-pooling operation is attached after the two convolution blocks. The feature map with a resolution of 2*2*2*128 is directly fed to the fully connected layers. 

\vspace{5pt}
\subsection{Hybrid Cascade Classifier Strategy}
\label{subsec: cascade_strategy}
In this paper, the cascade classifier network we propose is similar to~\cite{li2017not, huang2017multi-scale}. Unlike the weak classifier adopted in some cascade works~\cite{viola2001rapid, ouyang2017chained, shivajirao2019mask}, the cascade classifier at each stage in our method is actually a strong classifier, which only determines the objects with high confidence scores and leaves uncertain objects to next stage. When the top-1 score after softmax is higher than the threshold ${\theta^t}$ at stage $t$ ($t$=$1$ or $2$), the prediction of the sample at this stage is considered to be reliable, otherwise the sample will go through the next classifier and perform the comparison again until the last stage. 

Here, the threshold ${\theta^{t}}$ is not a constant value for different classes, since the difficulty of classifying objects with different classes varies (shown in Figure \ref{fig:cone_example}). We assume that ${\theta^{t}_{C}}$ is the threshold of class ${C}$ at the stage $t$. In order to maximize the inference speed and ensure the classification accuracy, the confidence $p^t$ (common to each class) at stage $t$ is introduced to adaptively determine the ${\theta^t_{C}}$. For each class ${C}$, the optimal ${\theta^t_{C}}^*$ in theory is expressed by

\begin{equation} 
{\theta^t_{C}}^* = \mathop{\arg\max}\limits_{\theta_{C}^t}({{N^+}^{t}_{C}(\theta_{C}^t)}),
\end{equation}

\begin{equation} 
s.t. \quad \frac{{N^+}^{t}_{C}(\theta_{C}^t)}{{N^+}^{t}_{C}(\theta_{C}^t)+{N^-}^{t}_{C}(\theta_{C}^t)}>=p^t
\end{equation}

\noindent where ${N^+}^{t}_{C}$ and ${N^-}^{t}_{C}$ denote the number of positive and negative samples of class $C$ whose top-1 scores are bigger than $\theta_{C}^t$, respectively. Given a specified confidence $p^t$, ${N^+}^{t}_{C}$ and ${N^-}^{t}_{C}$ should satisfy the above confidence constraint. However, in practice, we find that according to the threshold configuration of the above equations, the highest accuracy cannot always be achieved. This is because the negative sample distribution of each class for different classifiers is not the same. Therefore, simply throwing the negative samples of each class according to a common confidence standard to the subsequent classifier will decrease the accuracy. A revised standard is given by

\begin{equation} 
{\theta^t_{C}}^* = \mathop{\arg\max}\limits_{\theta_{C}^t}({{N^+}^{t}_{C}(\theta_{C}^t)}),
\end{equation} 

\begin{equation} 
s.t. \quad \frac{{N^+}^{t}_{C}(\theta_{C}^t) - {N^+}^{t}_{C}(\theta_{C}^t)|_{p^t=1.0}}{{N^+}^{t}_{C}(\theta_{C}^t) - {N^+}^{t}_{C}(\theta_{C}^t)|_{p^t=1.0} + {N^-}^{t}_{C}(\theta_{C}^t)}>=q^t
\end{equation}

\noindent where ${N^+}^{t}_{C}(\theta_{C}^t)|_{p^t=1.0}$ denotes the maximum number that all pass samples of each class are correct with a specific threshold $\theta_{C}^t|_{p^t=1.0}$. The number ${N^+}^{t}_{C}(\theta_{C}^t)-{N^+}^{t}_{C}(\theta_{C}^t)|_{p^t=1.0}$ represents the incremental changes of positive samples, as the threshold decreases. ${N^-}^{t}_{C}(\theta_{C}^t)$ represents the number of incremental changes of negative samples. $q^t$ denotes the incremental confidence which is more effective for improving accuracy. As a result, more samples will be classified at early stages, which can further improve the classification accuracy in later stages and speed up the inference time.  

Let $\eta_{C}^t$ be the top-1 score of a sample classified by class $C$ at the stage $t$. If the value $\eta_{C}^t$ is greater than $\theta_{C}^t$, this sample is then predicted credibly by the classifier at stage $t$. Otherwise, this sample will be fed to the next classifier. We denote that $\lambda^1$ and $\lambda^2$ are the overall pass rates of `easy' and `moderate' samples classified by stage 1 and stage 2 respectively in the whole test set. The mean inference time is given by:
\begin{equation}
T_{mean} = 1*T^{1} + (1-\lambda^1)*T^{2} + (1-\lambda^1-\lambda^2)*T^{3}
\end{equation}

\noindent where $T^i, i=1,2,3$ denotes the inference time of stage $i$ in Figure \ref{fig:cascade}. It is clear that, to achieve a fast inference time, the ratios of $\lambda^1$ and $\lambda^2$ should be larger as much as possible subject to the confidence constraints.

\vspace{5pt}

\section{Experiments}
\label{sec:expe}

\subsection{Datasets}

\noindent \textbf{Gray SDF.} The ModelNet40 dataset contains 12,311 CAD models from 40 classes and is split to 9843 training set and 2468 test set. Specifically, the ModelNet10 dataset is a subset of ModelNet40 dataset, containing 4899 CAD models from 10 categories.   
To obtain the SDF form of ModelNet40 and ModelNet10, first, we normalize each CAD model to a unit ball and transform it to a volumetric model through the solid voxelization method provided by PyMesh library~\cite{PyMesh}.
Then, we calculate the SDF of each voxel based on whether it is inside or outside the object by utilizing the C++/python library named edt~\cite{edt}.
The positive SDF means the distance from the exterior voxel grid to its closest surface while the negative SDF means the distance from the interior voxel grid to its closest surface. Before training, the SDF values are normalized to [-1, 1], according to the biggest absolute value.

\noindent \textbf{Colored SDF.} Colored SDF is consisted of three gray images standing for R, G, and B channels. One example has been illustrated in Figure \ref{fig:color_SDF}. Red channel image describes the interior of an object, while the blue channel image depicts the exterior of the object. Both of these parts are drawn with a single color. A gray SDF is stacked on the green channel image, where the gray value around the edge is set to 0 to enhance the visual appeal.

\noindent \textbf{3D MNIST.} The MNIST dataset~\cite{MNIST} contains 60000 training samples and 10000 test samples of hand-written digits from 0 to 9. The proposed 3D MNIST models are generated by stacking a number of 2D MNIST gray and SDF images in space, and then they are padded to $32^3$ with zero for the fair use of 3D CNN. In this paper, the numbers of stacked images are 4 and 28 for thin and thick 3D characters, respectively. The 3D MINIST dataset is used to verify the effectiveness of the proposed SDF voxel and the performance of the hybrid cascade network.

\subsection{Training Model of Cascade Network}

An overview of our approach has been illustrated in Figure \ref{fig:cascade}. Firstly, the FC-Net classifies simple objects very quickly at the first stage, with SDF values as input. Then, 3D CNNs with 2 and 4 convolution layers are adopted at the second and third stages respectively.

The experiments are carried out on an Intel 9900K CPU with 16G Memory and an NVIDIA 2080TI GPU. The software environments include PyTorch 1.2.0 and CUDA 10.

During the training process, classifiers in three stages are trained separately one-by-one. Once the early classifiers finish training, the network parameters in these parts will be fixed. In particular, FC-Net is completely independently trained without sharing any layer with the other two stages. The third stage shares part of convolution layers with the second stage. If the second network has completed training, the latter network no longer needs to train the parameters in the shared layers. 

The two confidence thresholds at each stage are treated as hyper-parameters and the optimal thresholds are determined by testing multiple threshold combinations on the test set. 


\subsection{Individual Classification of Three Stages}
\begin{figure}
\begin{center}
\includegraphics[width=0.48\textwidth]{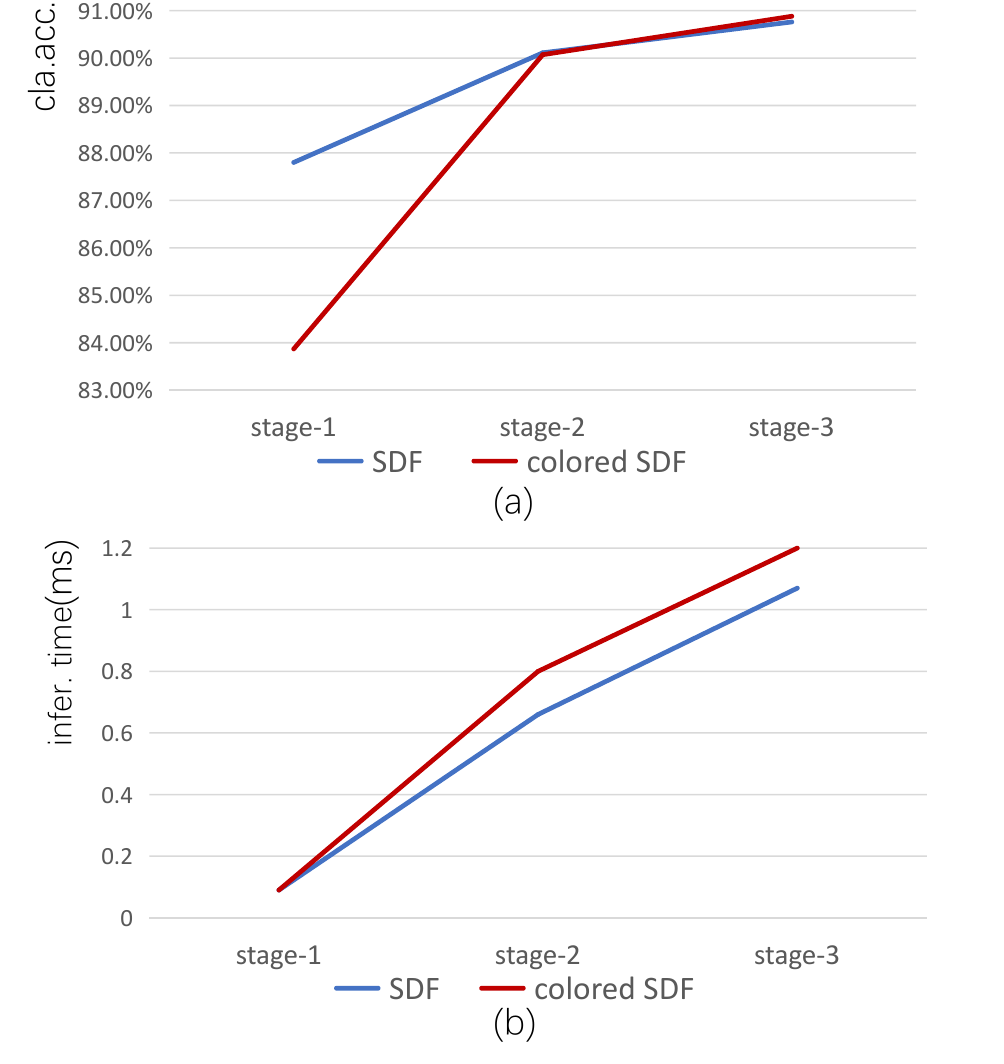}
\vspace{-12pt}
\caption{Accuracy and inference speed of the individual network at three stages with two kinds of SDF inputs. (a) accuracy results. (b) speed results.}
\label{fig:acc_speed_individual}
\end{center}
\vspace{-10pt}
\end{figure}

Before demonstrating the performance of the hybrid cascade method, we first test the performance of the individual network at each stage. Multiple SDF inputs are sent into the network to compare the performance differences. The results regarding the accuracy and running speed are depicted in Figure \ref{fig:acc_speed_individual}. Under the individual training, the network of the third stage can achieve the highest recognition accuracy among these three networks, but it costs the longest inference time. The performance of colored SDF voxel is worse than that of gray SDF voxel in the first-stage network, but the performances in the last two stages are similar. As a single-stage network, it is difficult to balance the fast inference speed and high accuracy simultaneously. Combining the advantages of different networks in a cascade style is expected to lift the overall performance.

\subsection{Cascade Classification}

In the proposed hybrid cascade network, the important parameters that affect final performance are confidences $p^1$, $p^2$, $q^1$, $q^2$. The middle part of Table~\ref{tab:pass_ratio} shows the numbers of pass and correct samples at stage 1 and 2 on ModelNet40 test set with different combinations of confidences. 
From Table~\ref{tab:pass_ratio}, it is clear that a trade-off have to be made between pass rates and accuracy of each stage. We finally choose $q^1= 0.8$, $q^2= 0.66$ in our cascade network as this combination gains the highest classification accuracy shown in the last row. 

The comparison of different approaches are shown in Table~\ref{tab:over_acc}. 
It can be seen that the overall accuracy of the proposed method with gray SDF voxel as input surpasses the SOTA voxel-based BPS-Conv3D \cite{prokudin2019efficient} which uses BPS grids with direction information as input, while consuming only about half of the inference time of BPS-Conv3D. Moreover, compared to the popular point cloud based methods, our result is also competitive with the coarse voxel input and much faster inference time.

\newcommand{\tabincell}[2]{\begin{tabular}{@{}#1@{}}#2\end{tabular}}
\begin{table*}
\caption{Performances on ModelNet40 test set under different combinations of confidences. The numbers of pass and correct samples at stage 1 and 2 are separately shown in the middle part. The overall classification accuracy is shown in the last row.}
\vspace{-10pt}
\label{tab:pass_ratio}
\begin{center}
\begin{tabular}{|l|c|c|c|c|c|c|}
\hline
\multirow{2}{*}{Confidence} &  $p^1= 1$ & $p^1= 0.98$  & $p^1= 0.96$  & $q^1= 0.9$ & $q^1= 0.8$  & $q^1= 0.8$ \\
& $p^2= 1$ & $p^2= 0.98$ & $p^2= 0.96$ & $q^2= 0.9$ & $q^2= 0.8$ & $q^2= 0.66$ \\
\hline
stage-1 & 1281/1281 & 1378/1368 & 1612/1572 & 1560/1549 & 1815/1770 & 1815/1770 \\
stage-2 & 440/440 & 403/401 & 327/324 & 408/402 & 316/300 & 473/414\\
\hline
Accuracy & 91.77\% & 91.65\% & 91.36\% & 91.97\% & 91.73\% & 92.01\%\\
\hline
\end{tabular}
\end{center}
\end{table*}

\begin{table*}[thp]
\caption{Accuracy, parameters, flops and inference time of different methods.}\label{tab:over_acc}
\vspace{-10pt}
\begin{center}
\begin{tabular}{|l|c|c|c|c|c|}
\hline
Method & Input signal & Accuracy (\%) & Params (M) & Flops & Infer. time (ms) \\
\hline
\hline
MVCNN-new(12x)~\cite{Su2015MultiviewCN} & images  & \textbf{95.0} & 130 & 9.0 x $10^{10}$ & 13.5\\
\hline
PointNet~\cite{Qi2017PointNetDL} &  xyz   & 89.2 & 3.5 & 4.4 x $10^8$ & 1.8\\
PointNet++~\cite{Qi2017PointNetDH} &  xyz   & 90.7 & 1.7 & 1.6 x $10^9$ & 2.9\\
DGCNN~\cite{Wang2018DynamicGC} &  xyz   & 92.2 & 1.8 & 2.5 x $10^9$ & 3.5\\
\hline
Voxnet~\cite{maturana2015voxnet} & binary voxel  & 83.0 & 0.9 & 6.0 x $10^7$ & \textbf{0.25}\\
OCNN~\cite{wang2017o-cnn} & octree voxel  & 89.9 & 1.3 & 2.1 x $10^8$ & 4.7\\
BPS-Conv3D\cite{prokudin2019efficient} & BPS grids+dire.  & 90.8 & 4.4 & 3.5 x $10^8$ & 0.55\\
\hline
Cascade network (\textbf{Ours}) & SDF voxel & 92.0 & 1.2 & 5.7 x $10^8$ & 0.30\\
\hline
\end{tabular}
\end{center}
\vspace{-10pt}
\end{table*}

\begin{figure*}[htbp]
\setlength{\belowcaptionskip}{-15pt}
\begin{center}
\includegraphics[width=0.99\textwidth]{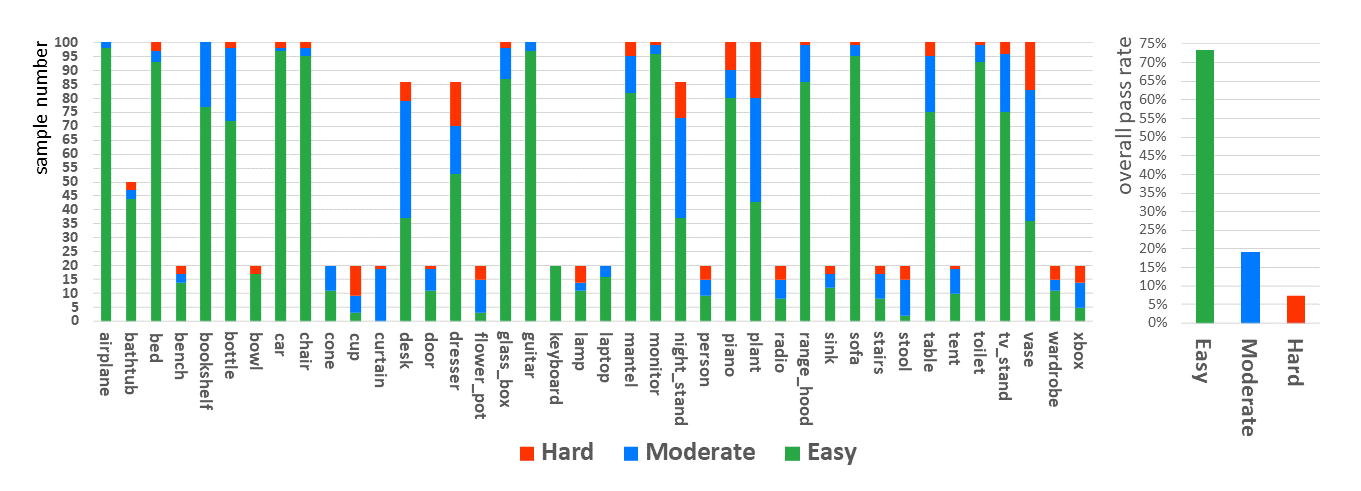}
\vspace{-10pt}
\caption{Two histograms of three categories. The left one plots sample numbers for each object class of ModelNet40. Different colors denote the difficulty types. The right one depicts the overall pass rate of each difficulty type, from which we can see that more than $70\%$ models can be recognized credibly in the first stage, whose network only consists of two hidden FC layers.}
\label{fig:cone_example}
\end{center}
\vspace{-10pt}
\end{figure*}

As illustrated in Fig.~\ref{fig:sink_example} and Fig.~\ref{fig:cascade}, the objects are divided into `easy', `moderate' and `hard' categories through the three-stage cascaded classifiers. Figure~\ref{fig:cone_example} shows the comprehensive results for each class of ModelNet40 test set. It is observed that the `easy' set accounting for the majority of all samples can be credibly predicted by the classifier in the first stage. Since an `easy' sample does not need to go through later stages containing 3D CNN layers, the mean computational cost is largely reduced. Moreover, the classification difficulty for different class can also be concluded from Figure~\ref{fig:cone_example}. For example, `plane' and `guitar' are the simplest among all shape classes, as $>\!95\%$ of their samples are classified by the first stage network and the residual samples are classified by the second stage network.

\subsection{Ablation Study}

The ablation experiments focus on what the performance is, when stage 1 or stage 2 is removed from the hybrid cascade architecture. Table~\ref{tab:ablation_removeOneStage} shows the results. With the removal of the first stage, the accuracy drops a little but the average inference time rises to 0.72ms. The reason is that stage 1 classifies 73.5\% test samples while the inference time only costs 0.09ms. Compared to stage 2 and 3, the inference time in stage 1 is much smaller. Besides, removing the second stage, the accuracy decreases compared with the removal of the first stage. The increase of inference time is not as much as the removal of the first stage. Overall, the three stages in the whole workflow are important in pursuing a system which runs fast, while a high accuracy is still kept.

\begin{table}
\caption{Ablation results of removing one stage.}\label{tab:ablation_removeOneStage}
\vspace{-10pt}
\begin{center}
\begin{tabular}{|l|c|c|}
\hline
Operation &  Accuracy (\%) & Infer. time (ms) \\
\hline
Remove stage-1 & 91.7  & 0.72 \\
Remove stage-2 & 91.5 & 0.37 \\
\hline
\end{tabular}
\end{center}
\vspace{-10pt}
\end{table}

\section{Further Investigation}
\label{sec:comparison}

\subsection{Influences from Different Input signals}

In this subsection, we further investigate influences of other aspects, including the binary voxel input, the low resolution voxel input and the performance on ModelNet10 data set.
The classification results are shown in Table~\ref{tab:3-1}. The detailed setup is same to Sec.~\ref{sec:expe}. The overall classification accuracy of ModelNet10 and ModelNet40 with vanilla $8^3$ binary voxel representation (shown in the first row of Table~\ref{tab:3-1}) is only $59.91\%$ and $55.27\%$, respectively. But we experimentally found that the classification accuracy can be greatly improved by $\sim\!30\%$ if using the SDF value as voxel input (shown in the second row of Table~\ref{tab:3-1}). This is probably due to the fact that FC-Net is sensitive to the proportion of the same input. When the input is converted to SDF, each voxel stores an distinguishable  signed distance to implicitly represent the local shape, just like each pixel in gray image has a distinguishable gray value. Thus the performance can be greatly improved. 
\begin{table}
\caption{Classification accuracy of FC-Net and CNN on ModelNet}
\label{tab:3-1}
\vspace{-10pt}
\begin{center}
\begin{tabular}{|l|c|c|}
\hline
Voxel Input &  ModelNet10 & ModelNet40 \\
\hline
 & \multicolumn{2}{c|}{FC-Net} \\
\hline
Binary($8^3$) & 59.91\% & 55.27\%\\
SDF($8^3$) & 92.40\% & 85.78\%\\
\hline
 & \multicolumn{2}{c|}{3D CNN} \\
\hline
Binary($64^3$) & 92.00\% & 87.20\%\\
SDF($64^3$) & 94.04\% & 90.76\%\\
\hline
\end{tabular}
\end{center}
\vspace{-10pt}
\end{table}

Another disadvantage of binary voxel representation is that its classification accuracy is obviously lower than that of surface point cloud. It is somewhat unreasonable because voxels already contain as much geometric information as point clouds. The third and fourth rows in Table~\ref{tab:3-1} show that, when we use SDF value instead of binary input of voxel, classification accuracy is obviously improved, and exceeds the accuracy of the classic PointNet (89.2\%) and PointNet++ (90.7\%) with point coordinates as input (see Table \ref{tab:over_acc}).

\subsection{Inspirations from MNIST Classification}

\begin{figure}[t]
\begin{center}
\includegraphics[width=0.48\textwidth]{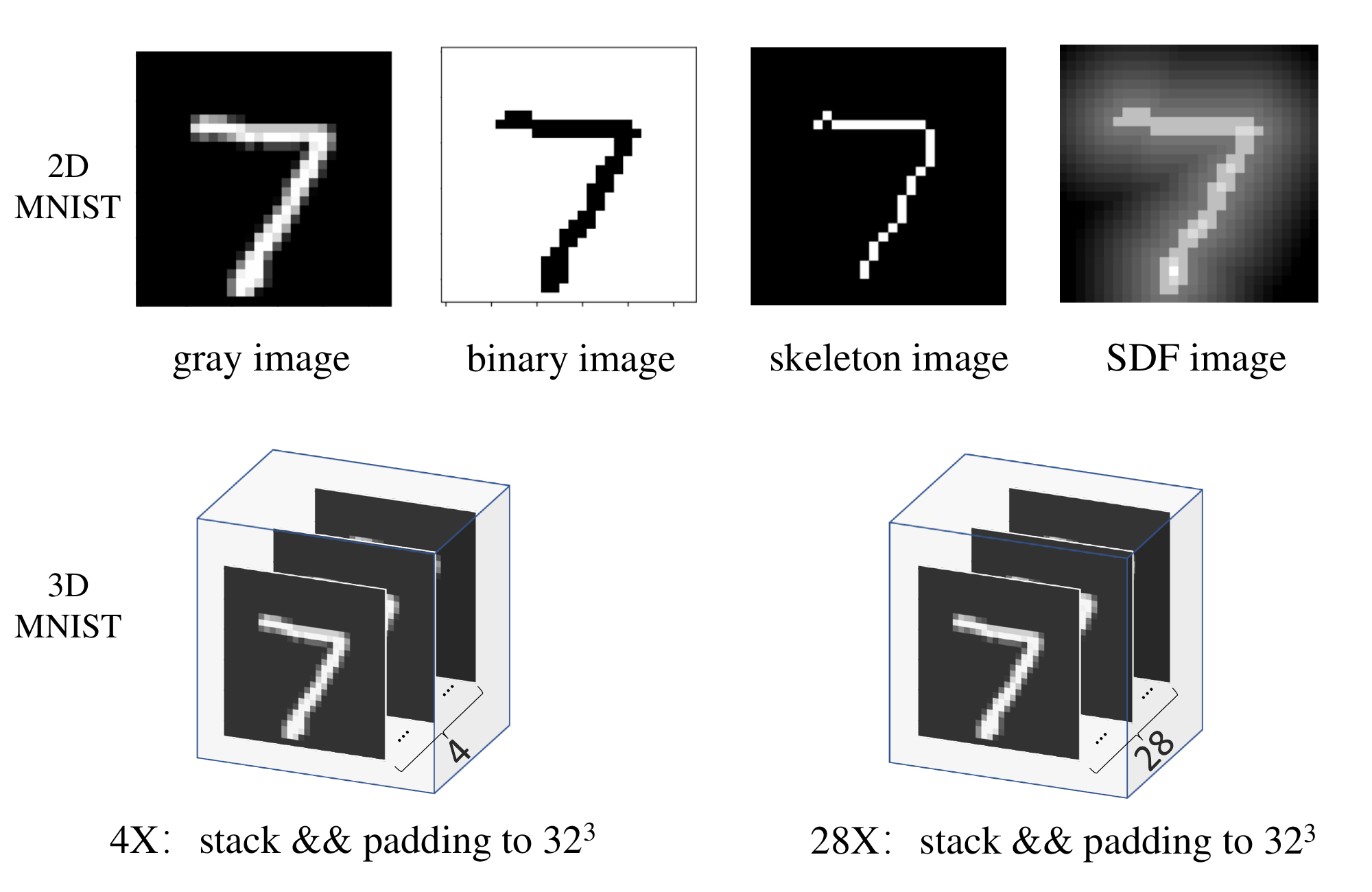}
\vspace{-12pt}
\caption{One example of different input signals of MNIST}\label{fig:MNIST_images}
\end{center}
\vspace{-10pt}
\end{figure}

\begin{figure}[t]
\setlength{\belowcaptionskip}{-10pt}
\begin{center}
\includegraphics[width=0.48\textwidth]{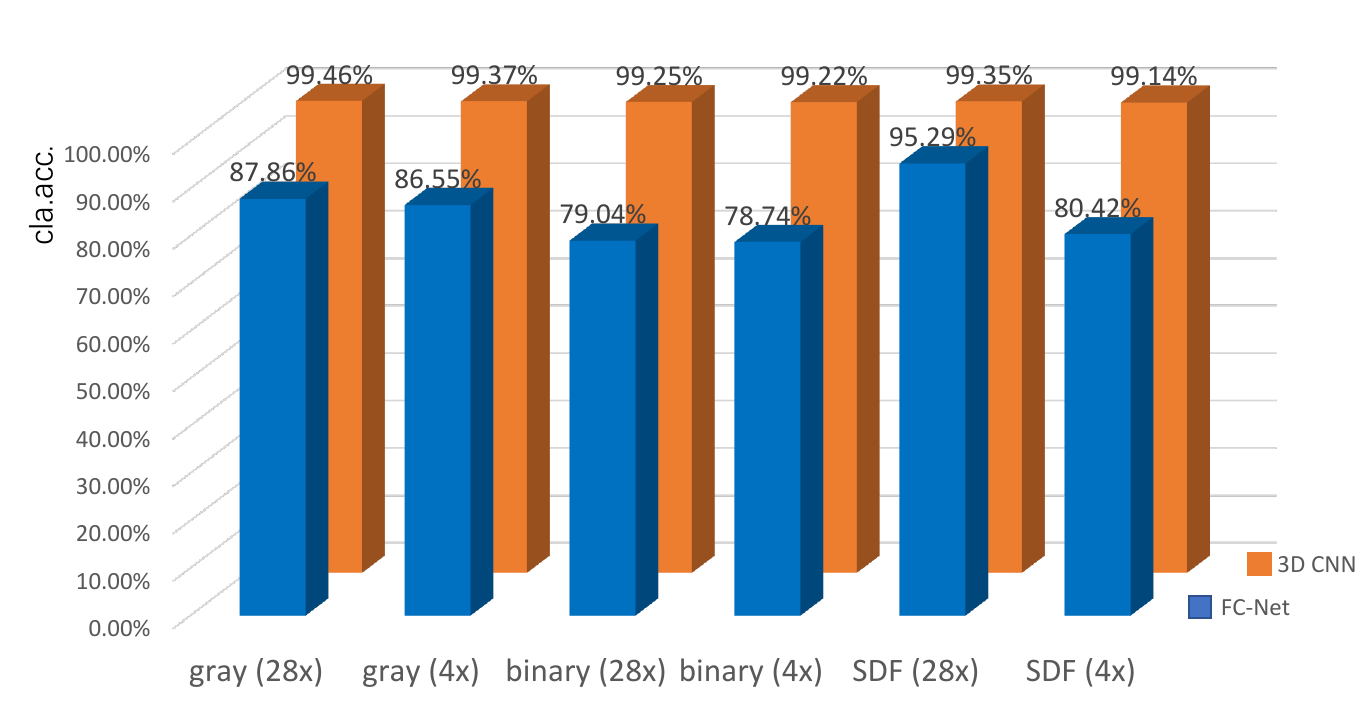}
\vspace{-10pt}
\caption{3D MNIST comparison of different voxel inputs.}
\label{fig:3D_MNIST_exp}
\end{center}
\vspace{-10pt}
\end{figure}

To further illustrate the influence of different input forms, we also conduct several MNIST classification experiments with different input signals. Besides the original gray image, there are some other variants of 2D images, e.g., binary image, silhouette image, SDF image. Figure  \ref{fig:MNIST_images} shows an example. As we have mentioned before, 3D MNIST models are generated through stacking multiple layers of the same image, followed by zero-padding. This process is repeated 4 and 28 times for  thin and thick 3D characters, respectively. 
The experimental results of 3D MNIST are shown in Figure \ref{fig:3D_MNIST_exp}. For thin 3D characters (4x), the classification accuracy of SDF voxel ranks only middle, slightly higher than the binary voxel. However, for thick 3D characters (28x), the accuracy of SDF voxel is significantly higher than other methods. This is because most of SDF values in 3D space of thin characters are almost the same, which limits the performance of FC-Net.


\section{Conclusions}
In this paper, we present a three-stage cascade network for 3D object classification. The accuracy of the proposed network exceeds all existing voxel-based methods and is close to the state-of-the-art point cloud based methods.
Meanwhile, the inference speed of this cascade network is also pretty fast. We owing the superior of our proposed cascaded network to several reasons. First, the network input we adopt is the SDF voxel instead of the traditional binary voxel. Second, the hybrid cascade classifier combines the advantages of FC-Net and 3D CNNs to achieve a better performance than an individual network. Third, the use of an adaptive threshold under the incremental confidence further improves the classification accuracy. 
Although the overall accuracy of the deeper network is better than that of the shallower network for some `hard' samples, the recognition speed of the shallower network is much fast than that of the deeper network. Overall, the proposed hybrid cascaded network has a satisfied performance both on accuracy and speed in 3D object classification.



{
\bibliographystyle{ieee}
\bibliography{ieee}
}

\end{document}